\documentclass{ifacconf}

\usepackage{graphicx}      
\usepackage{natbib}        
\usepackage{amssymb}  
\usepackage{amsmath}
\usepackage{times}
\usepackage{fancyhdr,graphicx}
\usepackage{algorithm}
\usepackage{algpseudocode}
\newcommand\norm[1]{\left\lVert#1\right\rVert}
\usepackage{multirow}
\usepackage{dblfloatfix} 

\begin{document}
\begin{frontmatter}

\title{Forward Dynamics Estimation from\\Data-Driven Inverse Dynamics Learning} 


\author[First]{Alberto Dalla Libera} 
\author[First]{Giulio Giacomuzzo} 
\author[First]{Ruggero Carli}
\author[Second]{Daniel Nikovski}
\author[Second]{Diego Romeres}

\address[First]{Department of Information Engineering, University of Padova, 
   via Gradenigo 6b, Padova, Italy (e-mail: alberto.dallalibera@unipd.it, giulio.giacomuzzo@phd.unipd.it, carlirug@dei.unipd.it).}
\address[Second]{Mitsubishi Electric Research Laboratories, 201 Broadway, Cambridge, MA, United States (e-mail: \{nikovski,romeres\}@merl.com)}

\begin{abstract}                
In this paper, we propose to estimate the forward dynamics equations of mechanical systems by learning  a model of the inverse dynamics and estimating individual dynamics components from it. We revisit the classical formulation of rigid body dynamics in order to extrapolate the physical dynamical components, such as inertial and gravitational components, from an inverse dynamics model. After estimating the dynamical components, the forward dynamics can be computed in closed form as a function of the learned inverse dynamics. We tested the proposed method with several machine learning models based on Gaussian Process Regression and compared them with the standard approach of learning the forward dynamics directly. Results on two simulated robotic manipulators, a PANDA Franka Emika and a UR10, show the effectiveness of the proposed method in learning the forward dynamics, both in terms of accuracy as well as in opening the possibility of using more structured~models.
\end{abstract}


\end{frontmatter}

\section{Introduction}
Methods for learning effectively the forward dynamics of mechanical systems, such as various robotic platforms, have been investigated for a long time. The forward dynamics describe the dynamical evolution of a system in response to forces, and are usually used for simulation and control purposes \cite{atkeson2002nonparametric,abbeel2006application,ng2006autonomous, romeres2019semiparametrical,dalla2020model,ota2021data}.
Knowledge of the forward dynamics is the basis of any model-based control algorithm, for example in model-based reinforcement learning (MBRL) \cite{amadio2022model,amadio2023learning}, where it is one element of a Markov decision process (MDP) formulation, generally called the transition function of the MDP, \cite{polydoros2017survey}.

The forward dynamics express the relationship between joint positions, joint velocities, applied forces, and joint accelerations. In RL terms, it represents the transition from the current state to the next state, given the current applied action. A model of the forward dynamics is often given by first-principles equations of motion, the so called rigid body dynamics (RBD). Physical models obtained from first principles are, however, often limited in their ability to describe some real physical phenomena, given our inability to describe certain complex dynamical behaviors of real systems. Moreover, the parameters of these models are often unknown or known only imprecisely. Consequently, in several applications, such models are not sufficiently accurate to describe the system dynamics. In this case, data collected by interacting with the system can be used to improve such models, leading to so called data-driven models.

Data-driven models have been widely studied in the context of inverse dynamics identification. The inverse dynamics is the inverse of the forward dynamics, and it takes as input the trajectory given in terms of joint positions, velocities, and accelerations, and  outputs the torques that are causing them. This function is often used in control schemes to improve, for example, the tracking performance of the controller \cite{craig2005introduction, nguyen2009model,dalla2021control}, see the survey \cite{Nguyen-Tuong2011} for an overview, and for anomaly detection see e.g., \cite{romeres2019anomaly}.

In the last decades, there has been an increased focus on learning inverse dynamics models of complex robotic systems by means of machine learning, using so called data-driven models, e.g. Gaussian processes regression (GPR) \cite{nguyen2008computed}, deep neural networks \cite{polydoros2015real}, support vector machines (SVM), etc. The critical aspect of data-driven solutions is generalization ability, i.e., the phenomenon that estimation accuracy might decrease in configurations that are far from the training samples. For this reason, several studies focused on deriving data-efficient estimators of the inverse dynamics, see, for instance, \cite{nguyen2009model, GIP, Rezaei_cascaded_GP} as black-box solutions. An alternative to black-box solutions are gray-box models, a combination of a physical model and a data-driven model: the physical models exploit prior knowledge, thus increasing data efficiency and generalization, whereas the data driven part compensates for the inaccuracies of the physical model, further improving accuracy. In the GPR literature, these models are named semiparametric models, see for instance, \cite{romeres2016online, romeres2019semiparametrical, SEMIPARAMTERIC_2016}. 


When solving the system identification problem for the dynamics of a mechanical system, the inverse dynamics have a significant advantage over the forward dynamics: the inverse dynamics model can be reformulated as a linear model in the inertial parameters \cite{hollerbach2008model}, whereas the model of the forward dynamics, in general, does not have such a formulation. This circumstance, while benefiting from the much more extensive set of techniques for learning linear models in comparison to nonlinear ones, also enjoys a structural advantage: learning the inverse dynamics is often better posed than learning the forward dynamics.


\textbf{Contributions.} In this work, we consider an alternative approach to learn the forward dynamics by taking into consideration the more favorable structural properties of the inverse dynamics. Instead of using the standard approach which learns a GP model directly on the forward dynamics, we propose to learn a data-driven inverse dynamics model by means of GPR, and then, compute the forwards dynamics with an exact and deterministic transformation from the learned inverse dynamics model. Experimental results show that, when compared to the standard approach, our strategy leads to better performance in terms of data efficiency and accuracy. Moreover, this strategy allows computing the forward dynamics with kernel types that were previously possible and specialized only to learn the inverse dynamics.

The paper is organized as follows. Section  \ref{sec:background} provides background formulation of dynamics models and GPR. The proposed approach is described in Section \ref{sec:proposed-approach}, while experiments are reported in Section \ref{sec:Experiments}. Section \ref{sec:Conclusions} draws the conclusions.


\section{BACKGROUND} \label{sec:background}
We start by providing the background formulation of the robot dynamics. Then, we describe GPR for inverse and forward dynamics identification, with details about the black-box priors adopted in this work.

\subsection{Rigid Body Dynamics}
Consider a mechanical system with $n$ degrees of freedom and denote with $\boldsymbol{q}_t \in \mathbb{R}^n$ its generalized coordinates at time $t$; $\dot{\boldsymbol{q}}_t$ and $\ddot{\boldsymbol{q}}_t$ are the velocity and the acceleration of the joints, respectively. The generalized torques, i.e., the control input of the system, are denoted by $\boldsymbol{\tau}_t \in \mathbb{R}^n$. For compactness, we will denote explicitly the dependencies on $t$ only when strictly necessary. Under rigid body assumptions, the dynamics equations of a mechanical system are described by the following matrix equation, called rigid body dynamics (RBD):
\begin{equation}\label{eq:dyn_eq}
    B(\boldsymbol{q}) \ddot{\boldsymbol{q}} + \boldsymbol{c}(\boldsymbol{q}, \dot{\boldsymbol{q}}) + \boldsymbol{g}(\boldsymbol{q}) + \boldsymbol{F}(\dot{\boldsymbol{q}})  = \boldsymbol{\tau} \text{,}
\end{equation}
where $B(\boldsymbol{q})$ is the inertia matrix, while $\boldsymbol{c}(\boldsymbol{q}, \dot{\boldsymbol{q}})$, $\boldsymbol{g}(\boldsymbol{q})$, and $\boldsymbol{F}(\dot{\boldsymbol{q}})$ account for the contributions of fictitious forces, gravity, and friction, respectively, see \cite{siciliano} for a more detailed description. For compactness, we introduce also $\boldsymbol{n}(\boldsymbol{q}, \dot{\boldsymbol{q}}) = \boldsymbol{c}(\boldsymbol{q}, \dot{\boldsymbol{q}}) +\boldsymbol{g}(\boldsymbol{q})+\boldsymbol{F}(\dot{\boldsymbol{q}})$, and the symbols $\hat{B}(\boldsymbol{q})$ and $\hat{\boldsymbol{n}}(\boldsymbol{q}, \dot{\boldsymbol{q}})$ will denote the estimates of $B(\boldsymbol{q})$ and $\boldsymbol{n}(\boldsymbol{q}, \dot{\boldsymbol{q}})$.
\subsubsection{Linear Model of Inverse Dynamics}
The model in eq.~\eqref{eq:dyn_eq} is linear w.r.t. the dynamics parameters, i.e., mass, center of mass, inertia, and friction coefficients of the links, see \cite{siciliano}. When neglecting friction, the number of dynamics parameters of each link is $p=10$, one for mass, three for center of mass, and six for the inertia tensor. Let $\boldsymbol{w} \in \mathbb{R}^{n \cdot p}$ be the vector collecting the dynamics parameters of all the links, then \begin{equation}\label{eq:dyn_eq_lin}
    \boldsymbol{\tau} = \Phi\left(\boldsymbol{q},\boldsymbol{\dot{q}},\boldsymbol{\ddot{q}}\right)\boldsymbol{w} = 
    \begin{bmatrix}
    \boldsymbol{\phi}^{(1)}\left(\boldsymbol{q},\boldsymbol{\dot{q}},\boldsymbol{\ddot{q}}\right)\\
    \vdots\\
    \boldsymbol{\phi}^{(n)}\left(\boldsymbol{q},\boldsymbol{\dot{q}},\boldsymbol{\ddot{q}}\right)
    \end{bmatrix}\boldsymbol{w}\text{,}
\end{equation}
where $\Phi\left(\boldsymbol{q},\boldsymbol{\dot{q}},\boldsymbol{\ddot{q}}\right) \in \mathbb{R}^{n\times(n \cdot p)}$ depends only on the kinematics parameters of the robot, i.e., its geometry.
\subsubsection{The forward dynamics} is the map that, given the current position, velocity, and torque $\boldsymbol{q},\dot{\boldsymbol{q}},\boldsymbol{\tau}$, outputs the acceleration $\ddot{\boldsymbol{q}}$. This model is needed for simulation and prediction purposes.

Forward dynamics learning is known to be more complex than learning the inverse dynamics. This can be seen directly in the RBD model given by physics. From eq.~\eqref{eq:dyn_eq}, we have
\begin{equation}
\boldsymbol{\ddot{q}} = B^{-1} \left( \boldsymbol{q} \right)  \left( \boldsymbol{\tau} - C \left( \boldsymbol{q}, \boldsymbol{\dot{q}} \right) \boldsymbol{\dot{q}} - \boldsymbol{g} \left( \boldsymbol{q} \right) \right) \text{,}
\label{eq:forward_dyn}
\end{equation} 

Equation~\eqref{eq:forward_dyn} cannot be written, in general, as a linear function of the dynamics parameters, as it can be done for the inverse dynamics, and the presence of the inverse of the inertia matrix creates a nonlinear dependence on the parameters. This complexity would be encoded in the black-box map that a machine learning algorithm would attempt to describe, if trying to learn the forward dynamics directly, i.e. $\boldsymbol{q},\dot{\boldsymbol{q}},\boldsymbol{\tau}\longrightarrow \ddot{\boldsymbol{q}}$.

\subsection{GPR for Inverse Dynamics Identification}\label{subsec:GPR}
GPR provides a solid probabilistic framework to identify the inverse dynamics from data. Typically, in GPR, each joint torque is modeled by a distinct and independent GP. Consider an input/output data set $\mathcal{D} = \left\{\boldsymbol{y}^{(i)}, X \right\}$, where $\boldsymbol{y}^{(i)} \in \mathbb{R}^N$~is a vector containing $N$ measurements of $\tau^{(i)}$, the \emph{i}-th joint torque, while $X=\left\{\boldsymbol{x}_{t_1}\dots\boldsymbol{x}_{t_N}\right\}$; the vector $\boldsymbol{x}_{t}$ contains the position, velocity, and acceleration of the joints at time $t$, hereafter designated as GP input. The probabilistic model of $\mathcal{D}$ is
\begin{equation*}
    \boldsymbol{y}^{(i)} =
    \begin{bmatrix}
    f^{(i)}\left(\boldsymbol{x}_{t_1}\right) \\ \vdots \\f^{(i)}\left(\boldsymbol{x}_{t_N}\right)
    \end{bmatrix}
    + \begin{bmatrix}
    e^{(i)}_{t_1} \\ \vdots \\ e^{(i)}_{t_N}
    \end{bmatrix}
      = \boldsymbol{f}^{(i)}(X) + \boldsymbol{e}^{(i)} \text{,}
\end{equation*}
where $\boldsymbol{e}^{(i)}$ is i.i.d. Gaussian noise with standard deviation $\sigma_i$, while $f^{(i)}(\cdot)$ is an unknown function modeled a priori as a GP, namely, $f^{(i)}(\cdot) \sim N(m_{f^{i}}(X),\mathbb{K}^{(i)}(X,X))$. $m_{f^{i}}(X)$ denotes the prior mean, and, generally, it is assumed to be equal to zero when no prior knowledge is available. The covariance matrix $\mathbb{K}^{(i)}(X,X)$ is defined through a kernel function $k^{(i)}(\cdot, \cdot)$. Specifically, the covariance between $f^{(i)}\left(\boldsymbol{x}_{t_j}\right)$ and $f^{(i)}\left(\boldsymbol{x}_{t_l}\right)$, i.e., the element of $\mathbb{K}^{(i)}(X,X)$ at row \emph{j} and column \emph{l}, is equal to $k^{(i)}\big(\boldsymbol{x}_{t_j}, \boldsymbol{x}_{t_l}\big)$. Exploiting the properties of Gaussian distributions, it can be proven that the posterior distribution of $f^{(i)}$ given $\mathcal{D}$ in a general input location $\boldsymbol{x}_{*}$ is Gaussian, see  \cite{rasmussen_GP_for_ML} for a comprehensive description. Then, the maximum a posteriori estimator corresponds to the mean, which is given by the following expression
\begin{equation}\label{eq:GP_estimate}
    \hat{f}^{(i)}(\boldsymbol{x}_*) = \mathbb{K}^{(i)}\left(\boldsymbol{x}_*,X\right)\boldsymbol{\alpha}^{(i)} + m_{f^{i}}(\boldsymbol{x}_*) \text{,}
\end{equation}
where
\begin{align*}
    &\boldsymbol{\alpha}^{(i)} = (\mathbb{K}^{(i)}\left(X,X\right) + \sigma_i^2 I)^{-1}\left(\boldsymbol{y}^{(i)}-m_{f^{i}}(X)\right) \text{,}\\
    &\mathbb{K}^{(i)}\big(\boldsymbol{x}_*,X\big) = \left[k^{(i)}\big(\boldsymbol{x}_*, \boldsymbol{x}_{t_1}\big) \dots k^{(i)}\big(\boldsymbol{x}_*, \boldsymbol{x}_{t_N}\big)\right] \text{.}
\end{align*}

Different solutions proposed in the literature can be grouped roughly based on the definition of the GP prior. In this paper, we will consider two black-box approaches, where the prior is defined without exploiting prior information about the physical model, and assuming $m_{f^{i}}(X)=0$.

\subsubsection{Squared Exponential kernel} The Squared Exponential (SE) kernel defines the covariance between samples based on the distance between GP inputs, see, for instance,  \cite{rasmussen_GP_for_ML}, and it is defined by the following expression
\begin{equation}\label{eq:SE_kernel}
    k_{SE}\big(\boldsymbol{x}_{t_j}, \boldsymbol{x}_{t_l}\big) = \lambda e^{-\norm{\boldsymbol{x}_{t_j}-\boldsymbol{x}_{t_l}}^{2}_{\Sigma^{-1}}} \text{;}
\end{equation}
$\lambda$ and $\Sigma$ are kernel hyperparameters. The former is a positive scaling factor, and the latter is a positive definite matrix which defines the norm used to compute the distance between inputs. A common choice consists in considering $\Sigma$ to be diagonal, with the positive diagonal elements named lengthscales.

\subsubsection{Geometrically Inspired Polynomial kernel} The Geometrically Inspired Polynomial (GIP) kernel has been recently introduced in \cite{GIP}. This kernel is based on the property that the dynamics equations in \eqref{eq:dyn_eq} are a polynomial function in a proper transformation of the GP input, fully characterized only by the type of each joint. Specifically, $\boldsymbol{q}$ is mapped to $\tilde{\boldsymbol{q}}$, the vector composed by the concatenation of the components associated with a prismatic joint and the sines and cosines of the revolute coordinates. As proved in \cite{GIP}, the inverse dynamics in \eqref{eq:dyn_eq} is a polynomial function in $\ddot{\boldsymbol{q}}$, $\dot{\boldsymbol{q}}$ and $\tilde{\boldsymbol{q}}$, where the elements of  $\ddot{\boldsymbol{q}}$ have maximum relative degree of one, whereas the ones of $\dot{\boldsymbol{q}}$ and $\tilde{\boldsymbol{q}}$ have maximum relative degree of two. To exploit this property, the GIP kernel is defined through the sum and the product of different polynomial kernels (\cite{MPK}), hereafter denoted as $k_P^{(p)}(\cdot,\cdot)$, where $p$ is the degree of the polynomial kernel. In particular, we have
\begin{align}
    &k_{GIP}\big(\boldsymbol{x}_{t_j}, \boldsymbol{x}_{t_l}\big) = \label{eq:GIP_eq}\\ 
    &\left(k_P^{(1)}\big(\ddot{\boldsymbol{q}}_{t_j}, \ddot{\boldsymbol{q}}_{t_l}\big) + k_P^{(2)}\big(\dot{\boldsymbol{q}}_{t_j}, \dot{\boldsymbol{q}}_{t_l}\big)\right)
    k_Q\big(\tilde{\boldsymbol{q}}_{t_j}, \tilde{\boldsymbol{q}}_{t_l}\big) \text{,} \nonumber
\end{align}
where, in its turn, $k_Q$ is given by the product of polynomial kernels with degree two, see \cite{GIP} for all the details. In this way, the GIP kernel allows defining a regression problem in a finite-dimensional function space where \eqref{eq:dyn_eq} is contained, leading to better data efficiency in comparison with the SE kernel.
\begin{itemize}
    \item Define the mean $m_{f^{i}}$ equal to the \emph{i}-th output of \eqref{eq:dyn_eq} or \eqref{eq:dyn_eq_lin}. Typically, in this case, the covariance is defined through an SE kernel, which aims at compensating inaccuracies of the prior mean.
    \item Define $m_{f^{i}}=0$, and the kernel as the sum of two kernels. The first kernel is a linear kernel derived from \eqref{eq:dyn_eq_lin} assuming that $\boldsymbol{w}\sim N(0, \Sigma)$, whereas the second is a standard SE kernel. Then, $k^{(i)}_{SP}$, namely, the semiparametrical kernel of the \emph{i}-th joint, is defined by the following expression:
    \begin{equation*}\label{eq:SP_kernel}
        k^{(i)}_{SP}\big(\boldsymbol{x}_{t_j}, \boldsymbol{x}_{t_l}\big) = \phi^{(i)}\big(\boldsymbol{x}_{t_j}\big) \Sigma \phi^{(i)}\big(\boldsymbol{x}_{t_l}\big)^T  + k^{(i)}_{SE}\big(\boldsymbol{x}_{t_j}, \boldsymbol{x}_{t_l}\big) \text{,}
    \end{equation*}
    where matrix $\Sigma$ is a tunable hyperparameter, typically assumed diagonal, expressing the prior covariance of $\boldsymbol{w}$, \cite{romeres2016online, romeres2019semiparametrical, SEMIPARAMTERIC_2016}.
\end{itemize}

We remark that we limit our investigation to black-box solutions, since we want to test the proposed approach w.r.t. standard direct learning of the forward dynamics (which cannot use semiparametric kernels). However, we would like to stress that our approach is fully compatible wih any kernel function.

\subsection{GPR for forward dynamics identification}\label{subsec:GPR_forward}
The GPR framework presented for the inverse dynamics learning can be applied also to the forward dynamics. When considering the \emph{i}-th joint, the input of the GP is the vector containing $\boldsymbol{q}$, $\boldsymbol{\dot q}$ and $\boldsymbol{\tau}$, while the output is the \emph{i}-th component of $\boldsymbol{\ddot q}$. However, w.r.t. the inverse dynamics, the choices for the GP prior are limited. (i) The GIP kernel cannot be applied, since it is based on the assumption that $\boldsymbol{\tau}$ is a polynomial function in a proper transformation of $\boldsymbol{q}$, $\boldsymbol{\dot q}$, and $\boldsymbol{\ddot q}$, but there is not an equivalent property for $\boldsymbol{\ddot q}$. (ii) Due to the fact that there is not an equivalent relation \eqref{eq:dyn_eq_lin}, i.e., $\boldsymbol{\ddot q}$ are not linear w.r.t. dynamics parameters, in general it is not possible to formulate a semiparametric kernel. Then, the options commonly available are:
\begin{itemize}
    \item If no prior is available, assume $m_{f^i}=0$, and define the covariance a priori as a SE kernel (or any non-structured kernel) with GP input $(\boldsymbol{q}, \dot{\boldsymbol{q}}, \boldsymbol{\tau})$;
    \item In case that \eqref{eq:dyn_eq} is known, a so called residual model can be used. The prior knowledge can be exploited by defining the mean $m_{f^i}$ to be equal to the \emph{i}-th component of $M(\boldsymbol{q})^{-1}(\boldsymbol{\tau}-\boldsymbol{n}(\boldsymbol{q}, \boldsymbol{\dot q}))$, and the covariance through an SE kernel, to compensate for eventual inaccuracies of the forward physical model.
\end{itemize}

\begin{figure*}[!hb]
\centering
  \includegraphics[width=0.9\linewidth]{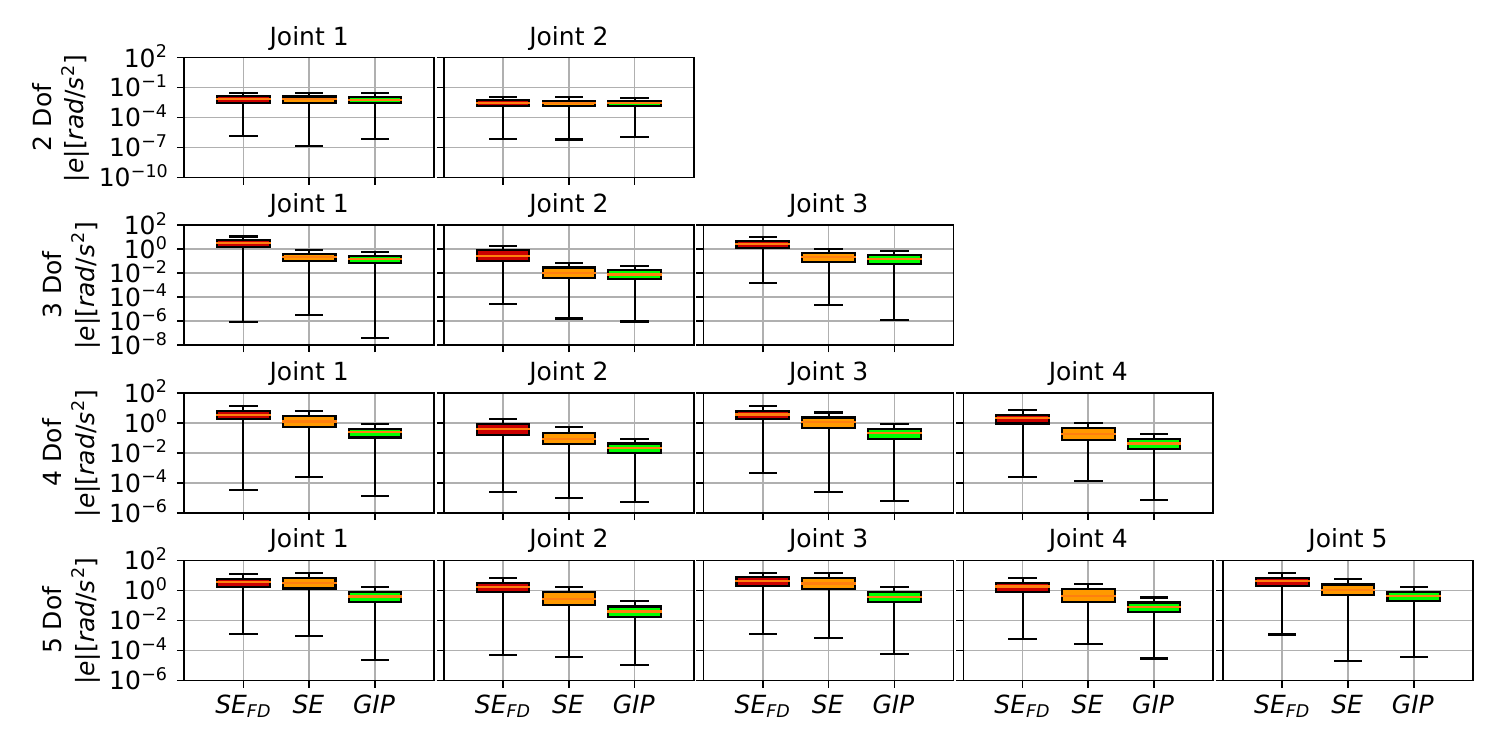}
  \caption{Distribution of the acceleration residual as a function of the DoFs on the simulated PANDA robot.}
  \label{fig:acc-residuals-distribution}
\end{figure*}

\section{Estimating Forward Dynamics from Inverse Dynamics Learning} \label{sec:proposed-approach}
In this section, we describe the proposed approach to estimating the forward dynamics \eqref{eq:forward_dyn} as a function of the inverse dynamics learned using GPR. In particular, we discuss how the physical equations of the RBD \eqref{eq:dyn_eq} entail an exact relationship that can be leveraged to compute gravitational contributions, inertial contributions, and $\boldsymbol{n}(\boldsymbol{q},\dot{\boldsymbol{q}})$ as a function of the inverse dynamics evaluated in a subset of the inputs. We assume that a distinct GP is used for each of the $n$ degrees of freedom, and we denote by $\hat{f}^{(i)}(\cdot)$, $i=1\dots n$ the estimator of the \emph{i}-th joint torque obtained by applying \eqref{eq:GP_estimate}. For convenience, from here on, we will point out explicitly the different components of the GP input, namely, the input of $\hat{f}^{(i)}$ will be $(\boldsymbol{q},\boldsymbol{\dot{q}},\boldsymbol{\ddot{q}})$ instead of $\boldsymbol{x}$, which comprises the concatenation of $\boldsymbol{q},\boldsymbol{\dot{q}},\boldsymbol{\ddot{q}}$. It is worth mentioning that the proposed approach is inspired by the strategy adopted in Newton-Euler algorithms, see \cite{deluca_NE_dyn_est}.

The proposed approach consists of (i) learning the inverse dynamics, which is a function suitable for identification purposes given its linear formulation, and (ii) using the learned model to have a closed-form deterministic transformation that estimates the forward dynamics. The method is described in Algorithm~\ref{alg}.

\textit{Assumption 1.}  The dependency of the dynamical components $B(\boldsymbol{q})\boldsymbol{\ddot{q}}$ and $\boldsymbol{n}(\boldsymbol{q},\dot{\boldsymbol{q}})$ w.r.t. the input quantities ${\boldsymbol{q},\dot{\boldsymbol{q}},\boldsymbol{\ddot{q}}}$ in \eqref{eq:dyn_eq} is exact. That is, the inertial contributions do not depend on $\dot{\boldsymbol{q}}$, the $\boldsymbol{n}(\boldsymbol{q},\dot{\boldsymbol{q}})$ do not depend on $\boldsymbol{\ddot{q}}$, and there are no terms with cross dependency that appear in the equations of motions.

Assumption 1 is a fairly mild assumption commonly assumed in classical manuscripts. The estimation of each dynamical component is described in the following.

In Algorithm \ref{alg}, initially the inverse dynamics, $\hat{f}^{(i)}$, are learned for each DoF of the mechanical system using GPR as described in Section~\ref{subsec:GPR}. These models are now considered known and fixed. Then, the inertia matrix $B \left( \boldsymbol{q} \right)$, the Coriolis and gravitational forces $\boldsymbol{n} \left( \boldsymbol{q}, \boldsymbol{\dot{q}} \right) = \boldsymbol{c} \left( \boldsymbol{q}, \boldsymbol{\dot{q}} \right) \boldsymbol{\dot{q}} + \boldsymbol{g} \left( \boldsymbol{q} \right)$ are estimated as a function of the learned inverse dynamics. Finally, the forward dynamics can be computed according to the physical laws \eqref{eq:forward_dyn}.

\begin{algorithm}[H]
\begin{algorithmic}
 \State collect data $\mathcal{D}=\{(\boldsymbol{q}_t,\boldsymbol{\dot{q}}_t,\boldsymbol{\ddot{q}_t}), \boldsymbol{\tau}_t \}$\;
 \State \textbf{Inverse Dynamics Learning}\\
 \For{$i=1...n_{DoF}$}
     $\hat{f}^i\leftarrow$ learn inverse dynamic, from \eqref{eq:GP_estimate}
  \EndFor
  \State\textbf{Forward Dynamics Estimation}\\
  \State $\hat{\boldsymbol{g}}(\boldsymbol{q})\leftarrow$ estimate gravitational component, from \eqref{eq:g_estimate}\\
  \State $\hat{B}(\boldsymbol{q})\leftarrow$ estimate inertial component, from \eqref{eq:B_ij_estimate}\\
  \State $\hat{\boldsymbol{n}}(\boldsymbol{q}, \dot{\boldsymbol{q}})\leftarrow$ estimate Coriolis and gravitaional component, from \eqref{eq:n_estimate}\\
  \State Compute $\boldsymbol{\ddot{q}} = \hat{B}^{-1} \left( \boldsymbol{q} \right)  \left( \boldsymbol{\tau} - \hat{\boldsymbol{n}} \left( \boldsymbol{q}, \boldsymbol{\dot{q}} \right) \right)$
 \caption{Inverse2Forward}
 \label{alg}
\end{algorithmic}
\end{algorithm}

\subsubsection{Gravitational contribution.} The motion equations in \eqref{eq:dyn_eq} show that the torque components due to the gravitational contributions account for all the terms that depend only on $\boldsymbol{q}$. Consequently, to obtain $g^{(i)}(\boldsymbol{q})$, i.e., the estimate of the \emph{i}-th gravitational contribution in the configuration $\boldsymbol{q}$, we evaluate $\hat{f}^{(i)}$ by setting $\boldsymbol{\dot{q}}=\boldsymbol{0}$, $\boldsymbol{\ddot{q}}=\boldsymbol{0}$. Then, the estimate of $\boldsymbol{g}(\boldsymbol{q})$ is 
\begin{equation}\label{eq:g_estimate}
    \hat{\boldsymbol{g}}(\boldsymbol{q}) =
    \begin{bmatrix}
    \hat{g}^{(1)}(\boldsymbol{q})\\
    \vdots\\
    \hat{g}^{(n)}(\boldsymbol{q})
    \end{bmatrix}
    =
    \begin{bmatrix}
    \hat{f}^{(1)}(\boldsymbol{q},\boldsymbol{0},\boldsymbol{0})\\
    \vdots\\
    \hat{f}^{(n)}(\boldsymbol{q},\boldsymbol{0},\boldsymbol{0})
    \end{bmatrix} \text{.}
\end{equation}

\subsubsection{Inertial contributions.} The inertial contributions, i.e., $B(\boldsymbol{q})\boldsymbol{\ddot{q}}$, account for all the contributions that depend simultaneously on $\boldsymbol{q}$ and $\boldsymbol{\ddot{q}}$. Consequently, to estimates these contributions, we evaluate the GP models in $(\boldsymbol{\ddot{q}}, \boldsymbol{0}, \boldsymbol{q})$, and subtract the gravitational contribution defined and computed previously. In particular, to obtain the element $\hat{B}_{ij}(\boldsymbol{q})$, i.e., the estimate of the $B(\boldsymbol{q})$ element in position $(i,j)$, we set all the accelerations to zero, except for the \emph{j}-th component. Denoting with $\boldsymbol{1}_j$ the vector with all elements equal to zero except for the $j$-th element, which equals one, we have
\begin{equation}\label{eq:B_ij_estimate}
   \hat{B}_{ij}(\boldsymbol{q}) = \hat{f}^{(i)}(\boldsymbol{q},\boldsymbol{0},\boldsymbol{1}_j) - \hat{g}^{(i)}(\boldsymbol{q}) \text{.}
\end{equation}

\subsubsection{Estimation of $\boldsymbol{n}(\boldsymbol{q},\dot{\boldsymbol{q}})$.} The vector $\boldsymbol{n}(\boldsymbol{q})$, defined under \eqref{eq:dyn_eq}, contains all the contributions that do not depend on $\ddot{\boldsymbol{q}}$. Then, $\hat{n}^{(i)}(\boldsymbol{q},\dot{\boldsymbol{q}})$, i.e., the estimate of the \emph{i}-th component of $\boldsymbol{n}(\boldsymbol{q})$, is computed by evaluating the \emph{i}-th GP model by setting $\ddot{\boldsymbol{q}}=\boldsymbol{0}$. Therefore, we can compute:
\begin{equation}\label{eq:n_estimate}
    \hat{\boldsymbol{n}}(\boldsymbol{q}, \dot{\boldsymbol{q}}) =
    \begin{bmatrix}
    \hat{n}^{(1)}(\boldsymbol{q}, \dot{\boldsymbol{q}})\\
    \vdots\\
    \hat{n}^{(n)}((\boldsymbol{q}, \dot{\boldsymbol{q}}))
    \end{bmatrix}
    =
    \begin{bmatrix}
    \hat{f}^{(1)}(\boldsymbol{q},\dot{\boldsymbol{q}},\boldsymbol{0})\\
    \vdots\\
    \hat{f}^{(n)}(\boldsymbol{q},\dot{\boldsymbol{q}},\boldsymbol{0})
    \end{bmatrix} \text{.}
\end{equation}

The proposed Algorithm~\ref{alg} is based on well-known first-principle relationships. Yet, it offers a useful bridge between modern machine learning techniques and the principles of classical mechanics that, as will be seen in Section~\ref{sec:Experiments}, improves significantly upon the standard approach to learn directly the forward dynamics.

A further advantage w.r.t. standard approaches is that more structured kernels can be utilized to learn the forward dynamics. As described in Sections~\ref{subsec:GPR}-\ref{subsec:GPR_forward}, the GIP and semiparametric kernels cannot be used to directly learn the forward dynamics. However, these kernels can be used to learn the inverse dynamics in the initial steps of Algorithm~\ref{alg}, and consequently used to learn the forward dynamics at the end of Algorithm~\ref{alg}.

Note that the method is not restricted to any function approximator to compute the inverse dynamics, e.g. neural networks could be used too; rather, it offers a general methodology, easy to implement, that reduces the problem of learning the forward dynamics to the more convenient inverse dynamics learning.


\section{Experiments}\label{sec:Experiments}
This section compares the learning performance of our approach with standard GP-based forward dynamics learning. Specifically, we implemented Algorithm \ref{alg} using as prior $m_{f^i}=0$ and the two black-box kernels introduced in \eqref{eq:SE_kernel} and \eqref{eq:GIP_eq}. The models obtained are hereafter referred to as $SE$ and $GIP$ estimators. As a baseline, we implemented the standard black-box forward dynamics estimator with $m_{f^i}=0$  and SE kernel described in Section \ref{subsec:GPR_forward}, hereafter denoted by $SE_{FD}$.

We carried out two experiments on simulated setups. The first investigated the behavior of the estimators as the DoFs increase on a simulated  PANDA Franka Emika robot. The second, instead, analyzed the data-efficiency performance on a UR10 robot. Training and test data sets of the kind: $\mathcal{D}=\{(\boldsymbol{q}_t,\boldsymbol{\dot{q}}_t,\boldsymbol{\ddot{q}_t}), \boldsymbol{\tau}_t \}$ are obtained by generating joint trajectories and evaluating \eqref{eq:dyn_eq} to compute the joint torques. The dynamics equations \eqref{eq:dyn_eq} are derived using the Python package \textit{Sympybotics}\footnote{https://github.com/cdsousa/SymPyBotics}, \cite{sympybotics}. All the estimators were implemented in Python, starting from \textit{gpr-pytorch}\footnote{https://bitbucket.org/AlbertoDallaLibera/gpr-pytorch}, a library for GPR based on \textit{pytorch}, \cite{pytorch}.

%
%

\subsection{Performance as a Function of DoFs}

\begin{figure}[!hb]
\centering
  \includegraphics[width=1\linewidth]{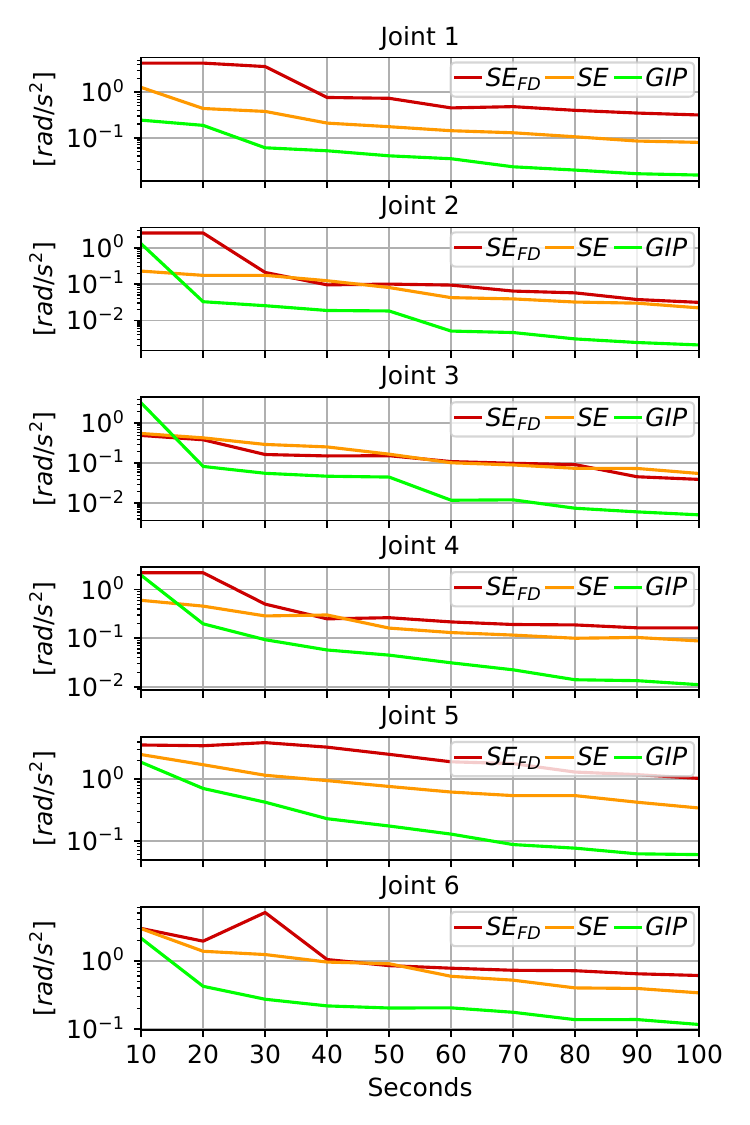}
  \caption{Median of the acceleration error modules as a function of the training data (in seconds), for each DoF.}
  \label{fig:data-efficiency}
\end{figure}

We compared the performance of the three estimators on a simulated PANDA robot as the number of DoFs increased from two to five. 
For each DoF, we collected a training and a test data set. The reference trajectories followed by the robot were different realization of Gaussian noise filtered with a low-pass filter, with cut-off frequency $1Hz$. The trajectories lasted $100$ seconds, collected at $100Hz$ for a total of $10000$ samples per data set. The joint torques were corrupted by Gaussian noise with standard  deviation $0.01Nm$. We computed the three models following Alg~\ref{alg} and trained the kernel hyperparameters by maximizing the marginal likelihood of the
training data, \cite{rasmussen_GP_for_ML}.\\
For each DoF, Figure \ref{fig:acc-residuals-distribution} reports the distribution of the acceleration error modules. It is particularly interesting to compare the performance of the $SE_{FD}$ and $SE$ estimators, since they adopt the same kernel to model, respectively, accelerations and torques. The proposed $SE$ estimator performs similarly to $SE_{FD}$ in the 2-DoF experiment, which is a simple test, and it ouptperforms the baseline in all the other experiments with higher DoFs. 
These results show that the proposed solution can improve the standard direct learning of the forward dynamics. \\
Interestingly, the $GIP$ estimator outperforms the baseline approach and the $SE$ estimator for all the DoFs and for all the joints. The performance gap between the $GIP$ and $SE$ estimators increases with the increase of the DoF. While the distributions of the estimation errors of $GIP$ and $SE$ are similar for two and three DoF,  $GIP$ significantly outperforms $SE$ for four and five DoF. The deterioration of the $SE$ performance is due to the limited generalization properties of the SE kernel, which affects the accuracy of the inverse dynamics model when the DoF increases. Table \ref{tab:RMSE-inv-dyn} reports the Root Mean Squared Errors of the torques estimates for each DoF tested. The RMSEs of the $SE$ estimator grow much rapidly with the DoF, compared with the $GIP$ estimator. Torque estimation errors are amplified in Algorithm \ref{alg}, limiting also the accuracy of the forward dynamics estimation.\\ 
\begin{table}
    \centering
\begin{tabular}{ |c|c|c|c|c| } 
\hline
Kernel & 2 DoF & 3 DoF & 4 DoF& 5 DoF \\
\hline
SE & 0.0101
&   0.0347
&  0.2282
&  0.2636
\\ 
\hline
GIP &  0.0016
&  0.0088
& 0.0465
& 0.0413
\\ 
\hline
\end{tabular}
    \caption{Inverse dynamics RMSEs in $Nm$ of the $SE$ and $GIP$ estimators as a function of the DoF.}
    \label{tab:RMSE-inv-dyn}
\end{table}
Thanks to the higher data efficiency and generalization of the GIP kernel (see the RMSEs in Table \ref{tab:RMSE-inv-dyn}), the $GIP$ estimator estimates accelerations accurately also in the setup with 5 DoF.  The performance of the $GIP$ estimator shows another advantage of the proposed approach, that is, the possibility of using data-efficient solutions proposed for inverse dynamics. As discussed in Section \ref{sec:background}, the forward dynamics in \eqref{eq:forward_dyn} does not admit convenient representation like the linear model in \eqref{eq:dyn_eq_lin} or the  GIP polynomial representation described in \cite{GIP}. These advantages can be further exploited by relying on the SP kernels described in Section \ref{subsec:GPR}. Using these kernels in simulated data would be unfair, as we have the exact knowledge of the physical model generating the data.
\subsection{Data Efficiency Performance}
We compared the data efficiency of the $SE_{FD}$, $SE$ and $GIP$ estimators by evaluating their accuracy as a function of the amount of training data. In this experiment, we considered a simulated UR10 robot. As in the previous experiment, the reference trajectories followed in the training and test data sets are distinct realizations of Gaussian noise filtered at $1Hz$. In Fig.~\ref{fig:data-efficiency}, we reported the medians of the acceleration error modules as a function of the number of seconds of training samples used. We used the first $10, 20, \dots, 100$ seconds of training data to derive the 3 estimators, after optimizing the kernel hyperparameters by maximizing the marginal likelihood of the training data. Then, we tested the derived estimators in the whole test data set.\\
The  $SE_{FD}$ and $SE$ estimators perform similarly for joints 2 and 3, but the $SE$ estimator outperforms $SE_{FD}$ for the other joints. This confirms the efficacy of the proposed method as a general method that, with the same model, outperforms the standard approach to learning the forward dynamics directly. Figure \ref{fig:data-efficiency} shows that the $GIP$ estimator significantly outperforms the other models, both in terms of accuracy and data efficiency. After around only $30s$ of data the GIP estimators performs like the other estimators after $100s$. The performance of the $GIP$ estimator confirms that the proposed approach can significantly improve the direct forward dynamics learning by exploiting the data-efficient solutions proposed for inverse dynamics, which are not available for forward dynamics models. 

\section{Conclusions}\label{sec:Conclusions}
In this work, we present a black-box GP-based strategy to learn the forward dynamics model. The proposed algorithm defines a prior on the inverse dynamics function instead of directly modeling the forward dynamics. Based on the learned inverse dynamics model, the algorithm computes individual dynamics components, i.e., inertial, gravitational, and Coriolis contributions, to estimate the forward dynamics. Experiments carried out in simulated environments show that the proposed strategy can be more accurate and data-efficient than directly learning accelerations in a black-box fashion. The advantages w.r.t. the standard approach are particularly relevant when considering data-efficient kernels, such as the GIP kernel, which are not available to model directly the forward dynamics.  The proposed method is general and can be adapted to any estimator, both black-box or gray-box. Next, we will test the approach both on data from real robots and make use of semiparametric kernels, and on MBRL algorithms as transition function.

\bibliography{references}             
                                                   







\end{document}